# Particle Filters in Robotics

**Sebastian Thrun**
Computer Science Department
Carnegie Mellon University
Pittsburgh, PA 15213
http://www.cs.cmu.edu/~thrun

## Abstract

In recent years, particle filters have solved several hard perceptual problems in robotics. Early successes of particle filters were limited to low-dimensional estimation problems, such as the problem of robot localization in environments with known maps. More recently, researchers have begun exploiting structural properties of robotic domains that have led to successful particle filter applications in spaces with as many as 100,000 dimensions. The fact that every model—no mater how detailed—fails to capture the full complexity of even the most simple robotic environments has lead to specific tricks and techniques essential for the success of particle filters in robotic domains. This article surveys some of these recent innovations, and provides pointers to in-depth articles on the use of particle filters in robotics.

## 1   INTRODUCTION

One of the key developments in robotics has been the adoption of probabilistic techniques. In the 1970s, the predominant paradigm in robotics was model-based. Most research at that time focused on planning and control problems under the assumption of fully modeled, deterministic robot and robot environments. This changed radically in the mid-1980s, when the paradigm shifted towards reactive techniques. Approaches such as Brooks's behavior-based architecture generated control directly in response to sensor measurements [4]. Rejections of models quickly became typical for this approach. Reactive techniques were arguable as limited as model-based ones, in that they replaced the unrealistic assumption of perfect models by an equally unrealistic one of perfect perception. Since the mid-1990s, robotics has shifted its focused towards techniques that utilize imperfect models and that incorporate imperfect sensor data. An important paradigm since the mid-1990s—whose origin can easily be traced back to the 1960s—is that of probabilistic robotics. Probabilistic robotics integrates imperfect models and and imperfect sensors through probabilistic laws, such as Bayes rule. Many recently fielded state-of-the-art robotic systems employ probabilistic techniques for perception [12, 46, 52]; some go even as far as using probabilistic techniques at all levels of perception and decision making [39].

This article focuses on particle filters and their role in robotics. Particle filters [9, 30, 40] comprise a broad family of sequential Monte Carlo algorithms for approximate inference in partially observable Markov chains (see [9] for an excellent overview on particle filters and applications). In robotics, early successes of particle filter implementations can be found in the area of robot localization, in which a robot's pose has to be recovered from sensor data [51]. Particle filters were able to solve two important, previously unsolved problems known as the global localization [2] and the kidnapped robot [14] problems, in which a robot has to recover its pose under global uncertainty. These advances have led to a critical increase in the robustness of mobile robots, and the localization problem with a given map is now widely considered to be solved. More recently, particle filters have been at the core of solutions to much higher dimensional robot problems. Prominent examples include the simultaneous localization and mapping problem [8, 27, 36, 45], which phrased as a state estimation problem involves a variable number of dimensions. A recent particle-filter algorithm known as FastSLAM [34] has been demonstrated to solve problems with more than 100,000 dimensions in real-time. Similar techniques have been developed for robustly tracking other moving entities, such as people in the proximity of a robot [35, 44].

However, the application of particle filters to robotics problems is not without caveats. A range of problems arise from the fact that no matter how detailed the probabilistic model—it will still be wrong, and in particular make false independence assumptions. In robotics, all models lack important state variables that systematically affect sensor and actuator noise. Probabilistic inference if further complicated by the fact that robot systems must make decisions in real-time. This prohibits, for example, the use of vanilla (exponential-time) particle filters in many perceptual problems.

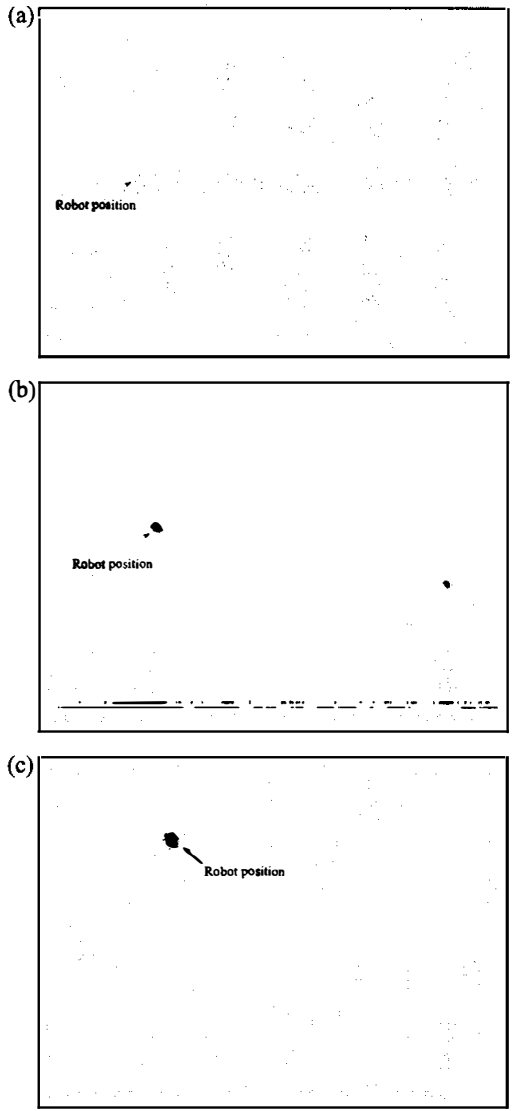

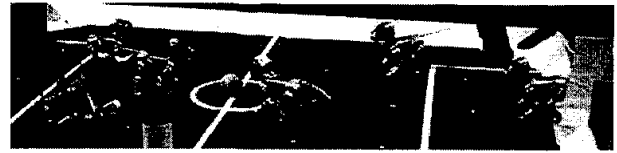

**Figure 2**: Particle filters have been used successfully for on-board localization of soccer-playing Aibo robots with as few as 50 particles [26].

**Figure 1**: Monte Carlo localization, a particle filter algorithm for state-of-the-art mobile robot localization. (a) Global uncertainty, (b) approximately bimodal uncertainty after navigating in the (symmetric) corridor, and (c) unimodal uncertainty after entering a uniquely-looking office.

This article surveys some of the recent developments, and points out some of the opportunities and pitfalls specific to robotic problem domains.

## 2   PARTICLE FILTERS

Particle filters are approximate techniques for calculating posteriors in partially observable controllable Markov chains with discrete time. Suppose the state of the Markov chain at time $t$ is given by $x_t$. Furthermore, the state $x_t$ depends on the previous state $x_{t-1}$ according to the probabilistic law $p(x_t \mid u_t, x_{t-1})$, where $u_t$ is the control asserted in the time interval $(t-1; t]$. The state in the Markov chain is not observable. Instead, one can measure $z_t$, which is a stochastic projection of the true state $x_t$ generated via the probabilistic law $p(z_t \mid x_t)$. Furthermore, the initial state $x_0$ is distributed according to some distribution $p(x_0)$. In robotics, $p(x_t \mid u_t, x_{t-1})$ is usually referred to as actuation or motion model, and $p(z_t \mid x_t)$ as measurement model. They as usually highly geometric and generalize classical robotics notions such as kinematics and dynamics by adding non-deterministic noise.

The classical problem in partially observable Markov chains is to recover a posterior distribution over the state $x_t$ at any time $t$, from all available sensor measurements $z^t = z_0, \ldots, z_t$ and controls $u^t = u_0, \ldots, u_t$. A solution to this problem is given by Bayes filters [20], which compute this posterior recursively:

$$p(x_t \mid z^t, u^t) \;=\; \text{const.} \;\cdot\; p(z_t|x_t) \int p(x_t|u_t, x_{t-1})$$
$$p(x_{t-1}|z^{t-1}, u^{t-1}) \; dx_{t-1} \qquad (1)$$

under the initial condition $p(x_0 \mid z^0, u^0) = p(x_0)$. If states, controls, and measurements are all discrete, the Markov chain is equivalent to hidden Markov models (HMM) [41] and (1) can be implemented exactly. Representing the posterior takes space exponential in the number of state features, though more efficient approximations exist that can exploit conditional independences that might exist in the model of the Markov chain [3].

In robotics, particle filters are usually applied in continuous state spaces. For continuous state spaces, closed form solutions for calculating (1) are only known for highly specialized cases. If $p(x_0)$ is Gaussian and $p(x_t \mid u_t, x_{t-1})$ and $p(z_t \mid x_t)$ are linear in its arguments with added independent Gaussian noise, (1) is equivalent to the Kalman filter [23, 32]. Kalman filters require $O(d^3)$ time for $d$-dimensional state spaces, although in many robotics problems the locality of sensor data allows for $O(d^2)$ implementations. A common approximation in non-linear non-Gaussian systems is to linearize the actuation and measurements models. If the linearization is obtained via a first-order Taylor series expansion, the result is known as extended Kalman filter, or EKF [32]. Unscented filters [21] obtain often a better linear model through (non-random) sampling. However, all these techniques are confined to cases where the Gaussian-linear assumption is a suitable approximation.

Particle filters address the more general case of (nearly) unconstrained Markov chains. The basic idea is to approxi-mate the posterior of a set of sample states $\{x_t^{[i]}\}$, or particles. Here each $x_t^{[i]}$ is a concrete state sample of index $i$, where is ranges from 1 to $M$, the size of the particle filter. The most basic version of particle filters is given by the following algorithm.

- **Initialization:** At time $t = 0$, draw $M$ particles according to $p(x_0)$. Call this set of particles $X_0$.

- **Recursion:** At time $t > 0$, generate a particle $x_t^{[i]}$ for each particle $x_{t-1}^{[i]} \in X_{t-1}$ by drawing from the actuation model $p(x_t \mid u_t, x_{t-1}^{[i]})$. Call the resulting set $\bar{X}_t$. Subsequently, draw $M$ particles from $\bar{X}_t$, so that each $x_t^{[i]} \in \bar{X}_t$ is drawn (with replacement) with a probability proportional to $p(z_t \mid x_t^{[i]})$. Call the resulting set of particles $X_t$.

In the limit as $M \to \infty$, this recursive procedure leads to particle sets $X_t$ that converge uniformly to the desired posterior $p(x_t \mid z^t, u^t)$, under some mild assumptions on the nature of the Markov chain.

Particle filters are attractive to roboticists for more than one reason. First and foremost, they can be applied to almost any probabilistic robot model that can be formulated as a Markov chain. Furthermore, particle filters are anytime [6, 53], that is, they do not require a fixed computation time; instead, their accuracy increases with the available computational resources. This makes them attractive to roboticists, who often face hard real-time constraints that have to be met using hard-to-control computer hardware. Finally, they are relatively easy to implement. The implementer does not have to linearize non-linear models, and worry about closed-form solutions of the conditional $p(x_t \mid u_t, x_{t-1})$, as would be the case in Kalman filters, for example. The main criticism of particle filter has been that in general, populating a $d$-dimensional space requires exponentially many particles in $d$. Most successful applications have therefore been confined to low-dimensional state spaces. The utilization of structure (e.g., conditional independences), present in many robotics problems, has only recently led to applications in higher dimensional spaces.

## 3  PARTICLE FILTERS IN LOW DIMENSIONAL SPACES

In robotics, the 'classical' successful example of particle filters is mobile robot localization. Mobile robot localization addresses the problem of estimation of a mobile robot's pose relative to a given map from sensor measurements and controls. The pose is typically specified by a two-dimensional Cartesian coordinate and the robot's rotational heading direction. The problem is known as position tracking if the error can be guaranteed to be small at all times [2]. More general is the the global localization problem, which is the problem of localizing a robot under global uncertainty. The most difficult variant of the localization, however, is the kidnapped robot problem [14], in which a well-localized robot is tele-ported to some other location without being told. This problem was reported, for example, in the context of the Robocup soccer competition [25], in which judges picked up robots at random occasions and placed them somewhere else [26]. Other localization problems involve multiple robots that can observe each other [15].

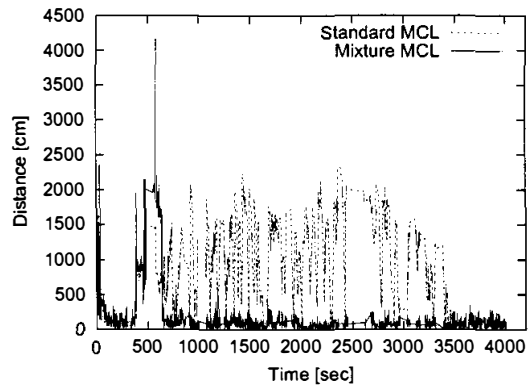

**Figure 3:** MCL with the standard proposal distribution (dashed curve) compared to MCL with a hybrid mixture distribution (solid line) [51]. Shown here is the error for a 4,000-second episode of camera-based MCL of a museum tour-guide robot operating in Smithsonian museum [50].

Figure 1 illustrates particle filters in the context of global localization of a robot in a known environment. Shown there is a progression of three situations, in which a number of particles approximate the posterior (1) at different stages of robot operation. Each particle is a sample of a three-dimensional pose variable, comprising the robot's Cartesian coordinates and its orientation relative to the map. The progression of snapshots in Figure 1 illustrate the development of the particle filter approximation over time, from global uncertainty to a well-localized robot.

In the context of localization, particle filters are commonly known as Monte Carlo localization (MCL) [7, 51]. MCL's original development was motivated by the condensation algorithm [19], a particle filter that enjoyed great popularity in computer vision applications. In most variants of the mobile localization problem, particle filters have been consistently found to outperform alternative techniques, including parametric probabilistic techniques such as the Kalman filter and more traditional techniques (see e.g., [18, 51]). MCL has been implemented with as few as 50 samples [26] on robots with extremely limited processing and highly inaccurate actuation, such as the soccer-playing AIBO robotic shown in Figure 2.

Recent research has led to a range of adaptations of the basic particle filter. Generating particles using the next state probability $p(x_t \mid u_t, x_{t-1})$ alone has been recognized as insufficient under a range of conditions, such as in the kidnapped robot problem [26] or in situations where the sensor accuracy is high in comparison to the control accuracy [51]. Common extensions involve hybrid sampling techniques, in which a subset of all samples is generated according to measurement model. Figure 3 shows a comparison of plain MCL versus an extended version using such a hybrid sampling scheme, obtained for data collected by a deployed mobile tour-guide robot. Other extensions regard the existence of unmodeled environmental state—which is a given in real-world robotics. A common approach inflates the uncertainty in the Markov chain model artificially, to accommodate systematic noise [16]. The same problem has also given rise to the development of probabilistic filters for pre-processing sensor data [16]. An example of the latter includes filters for range sensors that remove measurements corrupted by people [5]. MCL has also been made temporally more persistent by clustering particles and using different resampling rates for different clusters—a technique that empirically increases the robustness to errors in the map used for localization [33]. Stratified sampling techniques have been exploited to increase the variance of the resampling step [9]. Extensions to multi-robot localization problems are reported in [15].

## 4  PARTICLE FILTERS IN HIGH DIMENSIONAL SPACES

An often criticized limitation of plain particle filters is their poor performance in higher dimensional spaces. This is because the number of particles needed to populate a state space scales exponentially with the dimension of the state space, not unlike the scaling limitations of vanilla HMMs. However, many problems in robotics possess structure that can be exploited to develop more efficient particle filters.

One such problem is the simultaneous localization and mapping problem, or SLAM [8, 27, 36, 45]; see [48] for an overview. SLAM addresses the problem of building a map of the environment with a moving robot. The SLAM problem is challenging because errors in the robot's localization induce systematic errors in the localization of environmental features in the map. The absence of an initial map in the SLAM problem makes it impossible to localize the robot during mapping using algorithms like MCL. Furthermore, the robot faces a challenging data association problem of determining whether two environment features, observed at different point in time, correspond to the same physical feature in the environment. To make matters worse, the space of all maps often comprises hundreds of thousand of dimensions. In the beginning of mapping the size of the state space is usually unknown, so SLAM algorithms have to estimate the dimensionality of the problem as well. On top of all this, most applications of SLAM require real-time processing.

Until recently, the predominant approach for SLAM that meets most of the requirements above—with the exception of a sound solution of the data association problem—was based on extended Kalman filters, or EKFs [8, 36, 45]. As noted above, EKFs implement Equation (1) using linearized actuation and measurement models, with independent Gaussian noise. In their implementation, they crucially exploit that the environment is static. The resulting estimation problem is then described as a problem of jointly estimating a time-variant robot pose $x_t$ and the time-invariant location of $N$ features, denoted $y_1$ through $y_N$:

$$p(x_t, y_1, \ldots, y_N \mid z^t, u^t)$$
$$= \quad \text{const.} \cdot p(z_t | x_t) \int p(x_t | u_t, x_{t-1})$$
$$p(x_{t-1}, y_1, \ldots, y_N | z^{t-1}, u^{t-1}) \, dx_{t-1} \quad (2)$$

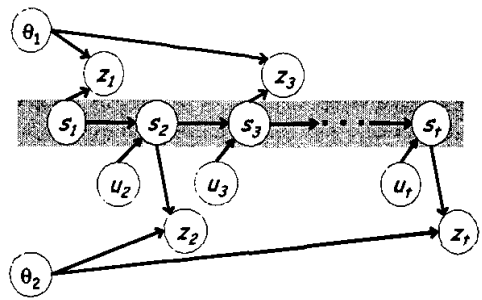

**Figure 4**: The SLAM problem as a 'dynamic Bayes network:' The robot moves from pose $x_1$ through a sequence of controls, $u_1, u_2, \ldots, u_t$. As it moves, it observes nearby features. At time $t = 1$, it observes feature $\theta_1$ out of two features, $\{\theta_1, \theta_2\}$. The measurement is denoted $z_1$ (range and bearing). At time $t = 1$, it observes the other feature, $\theta_2$, and at time $t = 3$, it observes $\theta_1$ again. The SLAM problem is concerned with estimating the locations of the features and the robot's path from the controls $u$ and the measurements $z$. The gray shading illustrates a conditional independence relation exploited by the FastSLAM algorithm.

Notice that this integration involves only the robot pose $x_t$, and not the variables $y_1 \ldots y_N$. For the linear-Gaussian model in EKFs, this integral is tractable [8, 45]. The EKF solution to the SLAM problem, however, suffers three key limitations:

1. First, the complexity of each update step is in $O(N^2)$, even in the absence of a data association problem. This limitation poses important scaling limitations. EKF algorithms can rarely manage more than a thousand features in realistic time. This limitation has spurred a flurry of research on hierarchical map representations, where maps are decomposed recursively into local submaps [17, 28]. Most of these approaches are still in $O(N^2)$ but with a constant factor that is orders of magnitude smaller than that of the monolithic EKF solution.

2. Second, EKFs cannot incorporate negative information, which is, they cannot use the fact that a robot failed to see a feature even when expected. The reason for this inability is that negative measurements give rise to non-Gaussian posteriors, which cannot be represented by EKFs.

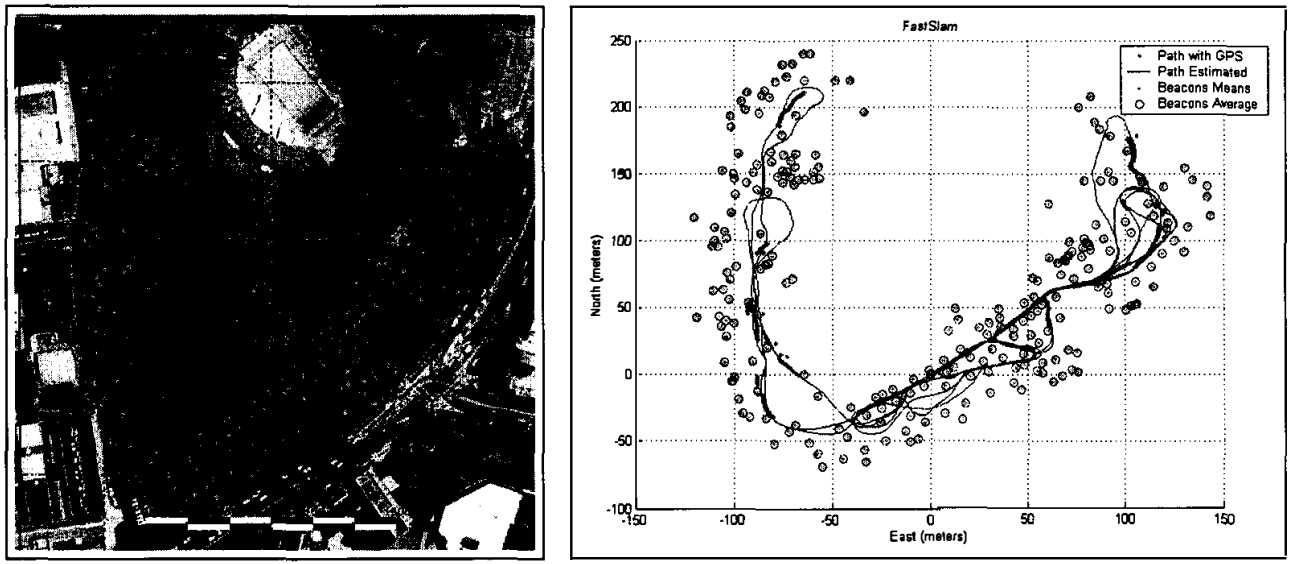

**Figure 5**: FastSLAM with real-world data: Shown here is a map of an outdoor environment (Victoria Park in Sydney), along with GPS information displayed here only for evaluating the accuracy of the resulting map. The resulting map error is extremely small, comparable in magnitude to the EKF solution. These results were obtained Juan Nieto, Eduardo Nebot, and Jose Guivant from the Australian Center of Field Robotics in Sydney, and are reprinted here with permission.

3. Third, EKFs provide no sound solution to the data association problem. It is common practice to use a maximum likelihood estimator for data association [8], using thresholding to detect (and reject) outliers. New environmental features are first placed on a provisional list, to reduce the odds of mistaking random noise for new, previously unseen features in the environment. However, as generally acknowledged in the literature [8], false data associations often lead to catastrophic failures.

The first and second of these limitations, and possibly the third, have recently been overcome by particle filters. Recent research [10, 34, 35, 37] has led to a family of so-called Rao-Blackwellized particle filters that, in the context ●f SLAM, lead to solutions that are significantly more efficient than the EKF. These particle filters require time $O(M \log N)$ instead of $O(N^2)$, where $M$ is the number of particles as above. Empirical evidence suggest that $M$ may be a constant in situations with bounded uncertainty—which includes all SLAM problems that can be solved via EKFs [34]. They can also incorporate negative information, hence make better use of measurement data than EKFs. Finally, highly preliminary experimental results suggest that particle filters provide a better solution to the data association problem than currently available with the EKF—although at present, this claim is not yet backed up by sufficient experimental evidence.

To understand particle filter solutions to the SLAM problem, it is helpful to analyze the structure of the SLAM problem. Assume, for a moment, that there is no data association problem, that is, the robot can uniquely identify individual features detected by its sensors. In this case, the SLAM problem is characterized by an important independence property [37], which is presently not exploited in EKF solutions. In particular, knowledge of the path of the robot renders the individual feature locations conditionally independent:

$$p(y_1, \ldots, y_N \mid x^t, z^t) \;=\; \prod_{n=1}^{N} p(y_n \mid x^t, z^t) \quad (3)$$

Figure 4 illustrates this independence, as explained in the caption to this figure. This important conditional independence property of the SLAM problem leads to the formulation of a more efficient version of Equation (2), one that estimates a posterior over robot paths $x^t$ (instead of poses $x_t$) along with the feature locations $y_n$:

$$p(x^t, y_1, \ldots, y_N \mid z^t, u^t)$$
$$= \; p(x^t \mid z^t, u^t) \, p(y_1, \ldots, y_N \mid x^t, z^t)$$
$$= \; p(x^t \mid z^t, u^t) \prod_{n=1}^{N} p(y_n \mid x^t, z^t) \quad (4)$$

A key property of particle filter is that each particle can be interpreted as a posterior over entire paths, and not just the present pose [9, 34]—a property that is not shared by EKFs. Thus, it is natural to implement the posterior over paths $p(x^t \mid z^t, u^t)$ in (4) by particle filters. The resulting feature estimators $p(y_n \mid x^t, z^t)$ are conditioned on individual particles representing path posteriors $p(x^t \mid z^t, u^t)$. However, since feature posteriors are conditionally independent given the path $x^t$, as stated in (3), the joint posterior over the features can be decomposed into separate estimators for each feature $p(y_n \mid x^t, z^t)$, for each $n = 1, \ldots, N$. The resulting filter maintains $M$ particles. Each particle contains a concrete robot path $x^t$, and a set of $N$ independent estimates of the locations of individual features in the environment.

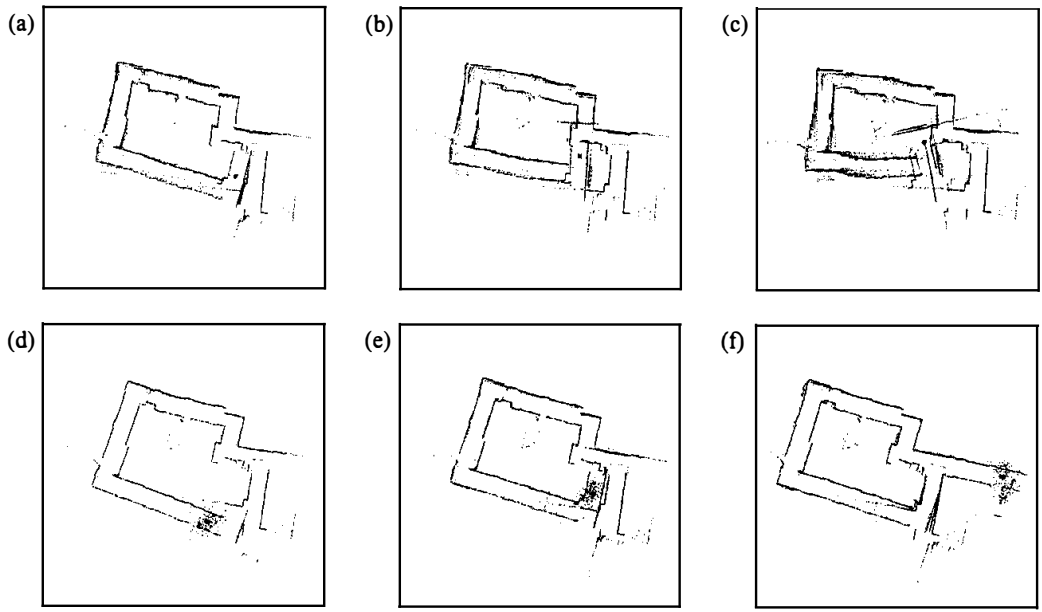

Figure 6: Lazy version of FastSLAM with unknown data association and point estimators (no variance) for environmental features: The top row (a) through (c) illustrates the result of mapping a cycle using conventional techniques. The bottom row (d) through (f) uses particle filters for estimating robot poses, hence is able to recover from large localization error that frequently occur when traversing cyclic environments. This real-time particle filter algorithm has proven to be highly robust in many environments.

In the FastSLAM algorithm [34], these feature posteriors $p(y_n \mid x^t, z^t)$ are implemented by EKFs, one for each feature. Each of these posteriors is of fixed dimension (e.g., 2 for landmarks in a plane). The resulting filter, thus, combines particle filters and Kalman filters: Posteriors over robot paths are represented by particle filters, very much as in the MCL algorithm described in the previous section. Each particle is then linked to $N$ Kalman filters, one for each feature in the map. Each of those particle filters corresponds to one feature, hence its dimension is independent of $N$. The resulting particle filter is called Rao-Blackwellized filter since the posterior for certain dimensions (the feature locations) are calculated exactly, whereas others (the robot pose) are approximated using particle filters.

Updating this Rao-Blackwellized particle filter in the naive way requires time $O(NM)$, where $M$ is the number of particles. Even the naive implementation avoids the quadratic complexity of the EKF solution by virtue of the decomposition in (3), which suggests that individual features can be localized independently when conditioned on a concrete robot path. The FastSLAM algorithm is even faster. Updates exploit the fact that the robot may only observe a finite number of features at any point in time. By representing feature estimates using tree structures as described in [34], the FastSLAM problem can be implemented in $O(M \log N)$ time. Initial empirical evidence in [34] sug-gest that under bounded robot pose uncertainty, this approach scales well even with constantly many particles $M$. This finding suggests that particle filters can (approximately) solve the SLAM problem in $O(\log N)$ time.

The use of particle filters opens the door to an improved solution to the data association problem. FastSLAM makes it possible to sample over data associations—rather than simply assuming that the most likely association is correct. Thus, FastSLAM implements a full Bayesian solution to the SLAM problem with unknown data association—something that has previously only been achieved using a recently developed mixture of Gaussian representation [11, 31]. FastSLAM can also incorporate negative information, that is, not seeing a feature that the robot expects to see. This is achieved by modifying the importance factors of individual particles accordingly.

As reported in [34], the Rao-Blackwellized particle filter algorithm delivers stat-of-the-art performance in large-scale SLAM problems, involving up to 100,000 dimensions. Figure 5 shows a typical result of FastSLAM obtained for an outdoor navigation problem. In this experiment, an autonomous land vehicle is used to map the location of trees in a public park.[1] This specific experiment involves several dozen circular features (stems of trees), which are detected using a laser range finder mounted on an outdoor vehicle. As the figure illustrates, the location of the trees and the vehicle is determined with high accuracy, which is comparable to the computationally much more cumbersome EKF solution. A second result is shown in Figure 6. This algorithm [49]—which was originally not stated as a version of FastSLAM—can be viewed as a lazy FastSLAM implementation where each particle uses maximum likelihood data association, and where feature locations are calculated in a lazy way, that is, for the most likely particle only. It also does not consider feature uncertainty in mapping, and instead memorizes the most likely feature location only. However, just like FastSLAM it uses particle filters to estimate posteriors over robot paths, and it uses those to find the most consistent map. This approach has been applied extensively to the problem of mapping indoor environments from laser range scans, where maps are collections of raw point features. It is presently one of the most robust real-time algorithms in existence for the indoor mapping problem with range finders.

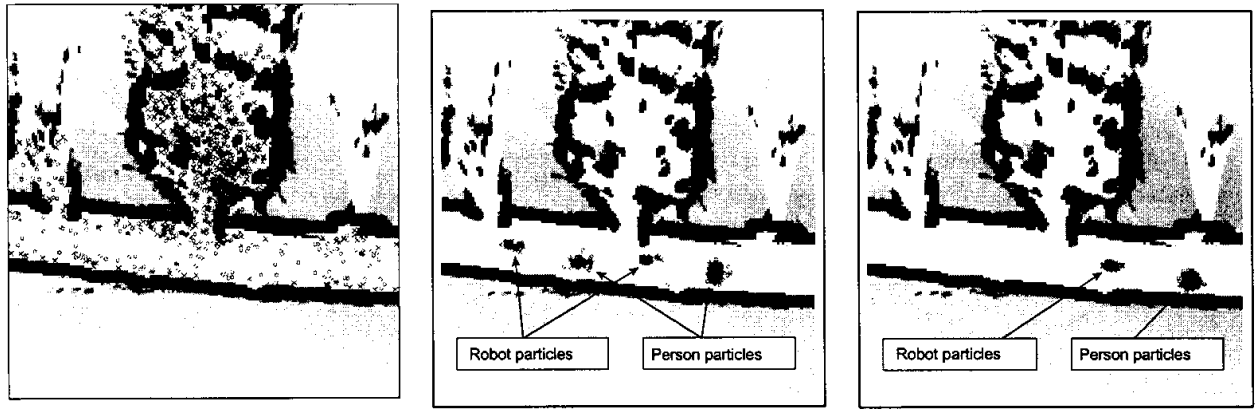

**Figure 7**: Particle filter-based people tracker: This algorithm uses maps to simultaneously localize a moving robot and an unknown number of nearby people. This sequence shows the evolution of the conditional particle filter from global uncertainty to successful localization and tracking of the robot.

In a similar vein, particle filters have also been used for tracking moving features, not just static ones. Here again, the underlying state spaces are high-dimensional, and particle filters have generated state-of-the-art results. In robotics, a prime example of tracking moving objects is the problem of tracking people [35, 44]. This problem is of great significance to the emerging field of service robotics [13, 43]. Service robots are robots designed to provide services to individual persons and hence have to be aware of where they are. The two dominant approaches that track people based on range measurements are both based on particle filters. The work by Schulz et al. [44] exploits a factored particle filter, where features are extracted from range measurements and associated with independent particle filters using maximum likelihood. The work by Montemerlo et al. [35] uses maps to detect people, in that it relies on differences between a previously acquired map of the environment and actual range scans, to identify and localize people. The map-based people tracking problem is similar in nature to the SLAM problem, since it involves the simultaneous localization of robots and people. The approach in [35] uses particle filters similar to FastSLAM, specifically exploiting the conditional independence prop-erty (3) that also applies to the people localization problem. Figure 7 depicts results using this particle filter-based people tracker under global initial uncertainty. The approach is able to simultaneously localize a robot under global uncertainty in real-time, and at the same time estimate the number and locations of nearby people. A similar approach for tracking the status of doors in mapped environments has been proposed in [1].

## 5   DISCUSSION

This article has described some of the recent successes of particle filters in the field of robotics. Traditionally, particle filters were mostly applied to low-dimensional robot localization problems-, where researchers have developed a rich repertoire of techniques to cope with the specifics of concrete robot environments. Recently, advanced variants of particle filters have provided new solutions to challenging higher-dimensional problems, such as the problem of robot mapping and people tracking. These approaches use hybrid representations that exploit structure in the underlying problems, expressed by conditional independences. In several of such structured robotics domains, particle filters are now among the most efficient and scalable solutions in existence.

Despite this progress, there exist plenty opportunities for future research. The most important opportunity concerns robot control: All the examples above address only the robot perception problem, but in robotics, the key problem is one of control: robots ought to do the right thing, no matter how their perception is organized. At present, relatively little is known about robot control under uncertainty. Recent developments in the field of partially observable Markov decision processes mostly address low-dimensional discrete spaces [22], and to the author's knowledge only a single algorithm exist that has applied continuous-space particle filters to such control problems [47]. This approach, however, is still too inefficient to be of relevance to robotics. Other discrete-state approximations have been developed [29, 42], yet they have only solved isolated navigation problems in mobile robotics.

More generally, recent research in AI has led to a great number of efficient algorithm for probabilistic inference in high-dimensional spaces with structure, beginning with the seminal work by Pearl [38]. Such techniques offer promising new solutions to hard robotics problems. The remaining challenge is to further develop them, and adapt them to the specific requirements characteristic of robotics domains.

## Acknowledgment

Much of the material presented here has been developed in close collaboration with the following people: Wolfram Burgard, Nando de Freitas, Frank Dellaert, Hugh Durrant-Whyte, Dieter Fox, Jose Guivant, Daphne Koller, Hannes Kruppa, Michael Montemerlo, Eduardo Nebot, Juan Nieto, Joelle Pineau, Nicholas Roy, and Ben Wegbreit. Their various contributions are gratefully acknowledged.

The research reported here is partially sponsored by DARPA's TMR Program (contract number DAAE07-98-C-L032), DARPA's MARS Program, and DARPA's CoABS Program (contract number F30602-98-2-0137). It is also sponsored by the National Science Foundation (regular grant number IIS-9877033 and CAREER grant number IIS-9876136), all of which is gratefully acknowledged.